%% file: paper.tex
\documentclass[sigconf]{aamas}

\usepackage{balance} % for balancing columns on the final page

\usepackage{hyperref}

\usepackage{listings}

\definecolor{codegreen}{rgb}{0,0.6,0}
\definecolor{codegray}{rgb}{0.5,0.5,0.5}
\definecolor{codepurple}{rgb}{0.58,0,0.82}
\definecolor{backcolour}{rgb}{0.95,0.95,0.92}
 
\lstdefinestyle{mystyle}{
    backgroundcolor=\color{backcolour},   
    commentstyle=\color{codegreen},
    keywordstyle=\color{magenta},
    numberstyle=\tiny\color{codegray},
    stringstyle=\color{codepurple},
    basicstyle=\footnotesize,
    breakatwhitespace=false,         
    breaklines=true,                 
    captionpos=b,                    
    keepspaces=true,                 
    numbers=left,                    
    numbersep=5pt,                  
    showspaces=false,                
    showstringspaces=false,
    showtabs=false,                  
    tabsize=2
}
 
\usepackage{physics}
\usepackage{tikz}
\usepackage{amsmath}
\usepackage{mathdots}
\usepackage{cancel}
\usepackage{color}
\usepackage{siunitx}
\usepackage{array}
\usepackage{multirow}
\usepackage{gensymb}
\usepackage{tabularx}
\usepackage{extarrows}
\usepackage{booktabs}
\usetikzlibrary{fadings}
\usetikzlibrary{patterns}
\usetikzlibrary{shadows.blur}
\usetikzlibrary{shapes}

\usepackage{balance} % for balancing columns on the final page
\usepackage{csquotes}
\newcommand{\probP}{\text{I\kern-0.15em P}}
\usepackage{etoolbox}
\patchcmd{\thebibliography}{\section*{\refname}}{}{}{}

\usepackage[T1]{fontenc}
\usepackage{graphicx}
\usepackage{color}

\usepackage[inline, shortlabels]{enumitem}
\usepackage{tabularx}
\usepackage{caption}
\usepackage{listings}
\usepackage{stfloats}
\usepackage{titlesec}
\usepackage{ragged2e}
\usepackage[linesnumbered,ruled,vlined]{algorithm2e}
\usepackage{float}
\usepackage[english]{babel}
\addto\extrasenglish{

}

\makeatletter
\gdef\@copyrightpermission{
  \begin{minipage}{0.2\columnwidth}
   \href{https://creativecommons.org/licenses/by/4.0/}{\includegraphics[width=0.90\textwidth]{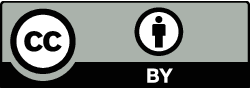}}
  \end{minipage}\hfill
  \begin{minipage}{0.8\columnwidth}
   \href{https://creativecommons.org/licenses/by/4.0/}{This work is licensed under a Creative Commons Attribution International 4.0 License.}
  \end{minipage}
  \vspace{5pt}
}
\makeatother

\setcopyright{ifaamas}
\acmConference[AAMAS '25]{Proc.\@ of the 24th International Conference
on Autonomous Agents and Multiagent Systems (AAMAS 2025)}{May 19 -- 23, 2025}
{Detroit, Michigan, USA}{Y.~Vorobeychik, S.~Das, A.~Nowé  (eds.)}
\copyrightyear{2025}
\acmYear{2025}
\acmDOI{}
\acmPrice{}
\acmISBN{}

\acmSubmissionID{605}

\title[AAMAS-2025 MOISE+MARL]{An Organizationally-Oriented Approach to Enhancing Explainability and Control in Multi-Agent Reinforcement Learning}

\author{Julien Soulé}
\affiliation{
  \institution{Univ. Grenoble Alpes}
  \city{Valence}
  \country{France}}
\email{julien.soule@lcis.grenoble-inp.fr}

\author{Jean-Paul Jamont}
\affiliation{
  \institution{Univ. Grenoble Alpes}
  \city{Valence}
  \country{France}}
\email{jean-paul.jamont@lcis.grenoble-inp.fr}

\author{Michel Occello}
\affiliation{
  \institution{Univ. Grenoble Alpes}
  \city{Valence}
  \country{France}}
\email{michel.occello@lcis.grenoble-inp.fr}

\author{Louis-Marie Traonouez}
\affiliation{
  \institution{Thales Land and Air Systems, BU IAS}
  \city{Rennes}
  \country{France}}
\email{louis-marie.traonouez@thalesgroup.com}

\author{Paul Théron}
\affiliation{
  \institution{AICA IWG}
  \city{La Guillermie}
  \country{France}}
\email{paul.theron@orange.fr}

\begin{abstract}
  Multi-Agent Reinforcement Learning can lead to the development of collaborative agent behaviors that show similarities with organizational concepts. Pushing forward this perspective, we introduce a novel framework that explicitly incorporates organizational roles and goals from the $\mathcal{M}OISE^+$ model into the MARL process, guiding agents to satisfy corresponding organizational constraints. By structuring training with roles and goals, we aim to enhance both the explainability and control of agent behaviors at the organizational level, whereas much of the literature primarily focuses on individual agents. Additionally, our framework includes a post-training analysis method to infer implicit roles and goals, offering insights into emergent agent behaviors. This framework has been applied across various MARL environments and algorithms, demonstrating coherence between predefined organizational specifications and those inferred from trained agents.
\end{abstract}


\newcommand{\BibTeX}{\rm B\kern-.05em{\sc i\kern-.025em b}\kern-.08em\TeX}

\begin{document}

%%% The following commands remove the headers in your paper. For final 
%%% papers, these will be inserted during the pagination process.

\pagestyle{fancy}
\fancyhead{}

%%% The next command prints the information defined in the preamble.

\maketitle 

%%%%%%%%%%%%%%%%%%%%%%%%%%%%%%%%%%%%%%%%%%%%%%%%%%%%%%%%%%%%%%%%%%%%%%%%

%%%%%%%%%%%%%%%%%%%%%%%%%%%%%%%%%%%%%%%%%%%%%%%%%%%%%%%%%%%%%%%%%%%%%%%%

%%% The acknowledgments section is defined using the "acks" environment
%%% (rather than an unnumbered section). The use of this environment 
%%% ensures the proper identification of the section in the article 
%%% metadata as well as the consistent spelling of the heading.

\section{Introduction}

% Context
Multi-Agent Reinforcement Learning (MARL) enables the discovery of a joint policy that controls agents' behaviors so they can achieve a global goal within a specific environment. 
This joint policy not only dictates the individual actions of agents but also manages their interactions with one another, and potentially with all other agents, without any preconceived notion of a predefined organization.

In environments that require social interaction among agents to optimally achieve the global goal, agents may converge in such a way that they exhibit recurring sets of similar behaviors across different testing episodes. 
These distinct sets of behaviors can demonstrate properties of specialization, complementarity, and stability, making them akin to implicit roles. Moreover, the trajectories of agents assuming these "implicit" roles may display similarities, such as recurrent observations at the end of each episode. These recurring patterns in agent histories can be interpreted as "implicit" goals, suggesting that agents may aim to pursue these as intermediate goals before reaching the global goal. These implicit roles and implicit goals form the foundation of an "implicit" structural and functional organization as defined in $\mathcal{M}OISE^+$~\cite{Hubner2007}.

However, it would be misleading to assume that all trained agents in any environment can be faithfully compared to a structural and functional organization. Indeed, we can interpret the behaviors of trained agents concerning their similarity to the potential vision of an implicit structural and functional organization, which we define as \textbf{organizational fit}.
While evaluating organizational fit would be useful to assess to what extent trained agents can naturally be explained as roles and goals, one could also consider the reverse approach. By guiding or encouraging agents to converge towards structural and functional organizations with higher organizational fit, we aim to enhance explainability and control in MARL.

\

% Problem
Building on these assumptions, this paper aims to further explore two key aspects:
\begin{enumerate*}[label={\roman*) },itemjoin={; \quad}]
    \item The \textbf{evaluation of organizational fit}, which seeks to measure how closely a joint policy aligns with a structural and functional organization. A significant challenge here is to understand under what conditions agents can be considered to form a structural and functional organization, given constraints imposed by the environment, goals, and other optional factors.
    Existing literature often addresses policy evaluation in terms of roles or goals~\cite{Isakov2024, Wen2024, Xie2024}, but these works generally lack a systematic and comprehensive approach. Current methods offer few clear tools for quantitatively and qualitatively measuring this organizational fit.
    \item The \textbf{control of organizational fit}, which aims to guide agents towards policies that conform to a structural and functional organization through user-defined constraints or incentives that implement roles and goals.
    The primary challenges include reducing the policy search space, improving convergence, and ensuring compliance with safety constraints.
    Existing approaches in this field often fall short in terms of enabling users to easily define and manage the application of organizational specifications in a practical manner within a standard MARL framework, without relying on paradigms such as Hierarchical Reinforcement Learning (HRL).
\end{enumerate*}

\

% Contribution
\noindent We introduce the \textbf{MOISE+MARL} framework, which integrates the Decentralized Partially Observable Markov Decision Process (Dec-POMDP) MARL framework with the $\mathcal{M}OISE^+$~\cite{Hubner2007} organizational model through proposed relationships. This framework allows users to manually define the logic of a role or a goal by relying on trajectory-based patterns to describe the expected behavior of an agent that has adopted a goal or mission. Once configured, they allow users to apply a role to an agent, adding constraints that automatically influence agents' policies by dynamically updating both the action space and reshaping the reward function. This framework also includes a method called \textbf{Trajectory-based Evaluation in MOISE+MARL} (TEMM), which uses unsupervised learning techniques to generalize implicit roles and implicit missions from observed trajectories across multiple test episodes. By measuring the gap between inferred implicit organizational specifications and actual behaviors, this method allows for a quantitative assessment of organizational fit. It is worth noting that unlike hierarchical reinforcement learning, which decomposes tasks into subtasks~\cite{Qi2024, Matsuyama2025, SaoMai2024}, our approach relies on explicit organizational roles and missions to guide agent coordination externally.

\

% Evaluation & Findings
We evaluated the MOISE+MARL framework in the following scenarios:
\begin{enumerate*}[label={\roman*) },itemjoin={; \quad}]
  \item Four distinct environments, each expected to result in the training of joint policies with different implicit organizations, to assess the generalizability of MOISE+ MARL's applicability
  \item Four MARL algorithms from the several families to assess their suitability with MOISE+ MARL during training and post-analysis
  \item Four sets of organizational specifications, one for each environment, to constrain agents in a manner that either enforces conformity intended for both manual and quantitative evaluation.
\end{enumerate*}

In all environments, we observed that agents having adopted roles do behave as expected according to their roles in a correlated way with a quantitative measure of the organizational fit by TEMM. The roles and missions inferred by TEMM closely align with the predefined specifications, demonstrating the internal consistency of MOISE+MARL, as the policy modifications introduced by organizational specifications are effectively captured by TEMM.
The results also indicate that policy-based and actor-critic algorithms are particularly well-suited for guiding agents towards stable policies. This stability allows agents to maintain consistent and coherent behaviors across episodes, which is essential for TEMM's generation of a stable implicit organization. In contrast, value-based algorithms showed greater variability in agent behaviors.

\

% Structure of the paper
\noindent The rest of the paper is organized as follows: \autoref{sec:related_works} presents works relative to evaluating and controlling organizational fit. \autoref{sec:moise_marl_framework} introduces the MOISE+MARL framework. \autoref{sec:TEMM_algorithm} describes the TEMM method. \autoref{sec:experimental_setup} describes the experimental protocol, particularly the environments and MARL algorithms. \autoref{sec:results} presents the experimental results. Finally, \autoref{sec:discussion_conclusion_future_work} discusses and concludes on the evaluation and control of organizational fit.

\section{Related works}
\label{sec:related_works}

This section explores works related to organizational fit, as framed by the two core issues introduced.

\subsection{Evaluating organizational fit}

Some works may be related to role or goal inference regarding the need to compute organizational fit or close concepts.
Wilson et al.~\cite{wilson2008learning} develop a method for transferring roles in Multi-Agent MDPs, which helps agents adapt by transferring roles across different environments. However, their model lacks the role abstraction as it focuses on specific, task-related roles.
Berenji and Vengerov~\cite{berenji2000learning} investigate coordination and role inference in UAV missions, enhancing cooperation through modeling agent dependencies. While useful for cooperation, their approach remains task-specific and does not provide the implicit role computation needed for organizational fit.
Yusuf and Baber~\cite{yusuf2020inferential} use inferential reasoning and Bayesian methods to facilitate task coordination among diverse agents. Though effective in dynamic coordination, their framework lacks role abstraction and does not measure alignment with an broader organizational structure either.
Serrino et al.~\cite{serrino2019finding} examine dynamic role inference in social settings, where agents deduce roles through interactions. While they enable flexible role understanding, their approach focuses on immediate operational roles rather than implicit roles that align with organizational models.

While some works explore organizational concepts in MARL, none explicitly address the computation of organizational alignment as we define it. Our concept of organizational fit requires a framework that assesses alignment with implicit goals.

\subsection{Controlling organizational fit}
Controlling organizational fit involves aligning the agents' policies with a predefined organization, often using constraints or incentives.
Achiam et al.~\cite{achiam2017cpo} introduce CPO, adjusting policies with safety constraints while maximizing rewards. MOISE+MARL, however, introduces constraints beyond safety to shape behavior toward organizational expectations by externally guiding agent learning.
Ray et al.~\cite{ray2019benchmarking} use Lagrange multipliers to integrate constraints into the reward function, balancing reward and constraint adherence. MOISE+MARL extends this by dynamically modifying the action space to enforce constraint adherence at various levels, offering flexible control over agent behaviors.
Safe exploration ensures agents learn while adhering to safety constraints. Garcia et al.~\cite{garcia2015comprehensive} overview methods for maintaining safe exploration, and Alshiekh et al.~\cite{alshiekh2018safe} propose shielding to block unsafe actions. MOISE+MARL goes further by using constraints to guide agents toward behaviors that align with organizational roles.
HRL breaks tasks into subtasks, aligning with organizational hierarchies. Ghavamzadeh et al.~\cite{ghavamzadeh2006hrl} illustrate that HRL can improve coordination. MOISE+MARL constrains MARL externally, offering a modular granularity and generating refined behaviors under organizational constraints.
Controlling Communication and Coordination is essential for ensuring organizational fit, especially in large-scale systems. Foerster et al.~\cite{foerster2018communication} propose decentralized coordination through shared knowledge, allowing agents to operate without centralized control.

Unlike HRL, the MOISE+MARL framework stands out for incorporating external organizational constraints that influence agents within a standard MARL framework, enabling modular granularity. Unlike Shielding or CPO, which typically focus on safety constraints, MOISE+MARL goes further by relying on actions and reward modifications to align with roles. % MOISE+MARL aims to handles scalability and adaptability by simplifying users interactions to defining and applying a smaller amount of organizational specifications.

\section{The MOISE+MARL framework}
\label{sec:moise_marl_framework}

This section introduces the formalism used to describe the functioning framework of the MOISE+MARL framework.

\subsection{Markov framework for MARL}

To apply MARL techniques, we rely on the \textit{Decentralized Partially Observable Markov Decision Process} (Dec-POMDP) \cite{Oliehoek2016}. Dec-POMDPs naturally model decentralized multi-agent coordination under partial observability, making them well suited for integrating organizational constraints. Unlike \textit{Partially Observable Stochastic Games} (POSG), the Dec-POMDP allows for a common reward function for agents, which promotes collaboration~\cite{Beynier2013}.

A Dec-POMDP $d \in D$ (where $D$ is the set of Dec-POMDPs) is defined as a 7-tuple $d = \langle S, \{A_i\}, T, R, \{\Omega_i\}, O, \gamma \rangle$, where $S = \{s_1,\dots,s_{|S|}\}$ is the set of possible states; $A_{i} = \{a_{1}^{i},\dots,a_{|A_{i}|}^{i}\}$ is the set of possible actions for agent $i$; $T$ represents the set of transition probabilities, with $T(s,a,s') = \probP(s'|s,a)$ as the probability of transitioning from state $s$ to state $s'$ following action $a$; $R: S \times A \times S \rightarrow \mathbb{R}$ is the reward function, assigning a reward based on the initial state, the action taken, and the resulting state; $\Omega_{i} = \{o_{1}^{i},\dots,o_{|\Omega_{i}|}^{i}\}$ is the set of possible observations for agent $i$; $O$ represents the set of observation probabilities, where $O(s',a,o) = \probP(o|s',a)$ is the probability of obtaining observation $o$ after performing action $a$ and reaching state $s'$; and $\gamma \in [0,1]$ is the discount factor
%, used to weight future rewards.

The following formalism is used with MOISE+MARL to solve the Dec-POMDP~\cite{Beynier2013,Albrecht2024}: $\mathcal{A}$ represents the set of $n$ \textbf{agents}; $\Pi$ denotes the set of \textbf{policies}, where a policy $\pi \in \Pi, \pi: \Omega \rightarrow A$ deterministically maps an observation to an action, representing the agent's internal strategy; $\Pi_{joint}$ represents the set of \textbf{joint policies}, with a joint policy $\pi_{joint} \in \Pi_{joint}, \pi_{joint}: \Omega^n \rightarrow A^n = \Pi^n$, which selects an action for each agent based on their respective observations, acting as a collection of policies used by agents within a team; $H$ is the set of \textbf{histories}, where a history (or trajectory) over $z \in \mathbb{N}$ steps (typically the maximum number of steps in an episode) is represented as the $z$-tuple $h = \langle \langle \omega_{k}, a_{k}\rangle | k \leq z, \omega \in \Omega, a \in A\rangle$, capturing successive observations and actions; $H_{joint}$ stands for the set of \textbf{joint histories}, with a joint history $h_{joint} \in H_{joint}$ over $z$ steps defined as the set of agent histories: $h_{joint} = \{h_1, h_2, \dots, h_n\}$; and finally, $V_{joint}(\pi_{joint}): \Pi_{joint} \rightarrow \mathbb{R}$ denotes the \textbf{expected cumulative reward} over a finite horizon (assuming $\gamma < 1$ or if the number of steps in an episode is finite), where $\pi_{joint}$ represents the joint policy for team $i$, with $\pi_{joint,-i}$ being the joint policies of other teams, considered as fixed.

% We refer to \textbf{solving the Dec-POMDP} as the search for a joint policy $\pi_{joint} \in \Pi_{joint}$ such that $\pi_{joint}s)$, achieving at least an expected cumulative reward of $s$, where $s \in \mathbb{R}$.

\subsection{The $\mathcal{M}OISE^+$ organizational model}

\begin{figure}[h!]
    \input{figures/moise_model.tex}
    \caption{A synthetic view of the $\mathcal{M}OISE^+$ model}
    \label{fig:moise_model}
\end{figure}

As illustrated in \autoref{fig:moise_model}, $\mathcal{M}OISE^+$ comprises three types of organizational specifications:

\noindent \paragraph{\textbf{Structural Specifications (SS)}} define how agents are structured, expressed as $\mathcal{SS} = \langle \mathcal{R}, \mathcal{IR}, \mathcal{G} \rangle$. $\mathcal{R}_{ss}$ is the set of roles ($\rho \in \mathcal{R}$) with an inheritance relation $\mathcal{IR}$ where $\rho_1 \sqsubset \rho_2$ if $\rho_1$ inherits from $\rho_2$. $\mathcal{GR}$ includes groups $\langle \mathcal{R}, \mathcal{SG}, \mathcal{L}^{intra}, \mathcal{L}^{inter}, \allowbreak \mathcal{C}^{intra}, \mathcal{C}^{inter}, np, ng \rangle$. Links ($\mathcal{L}$) define connections between roles: acquaintance, communication, or authority. Compatibilities $\mathcal{C}$ denote roles that agents can play together. Intra- and inter-group links and compatibilities are shown by $\mathcal{L}^{intra}$, $\mathcal{L}^{inter}$, $\mathcal{C}^{intra}$, and $\mathcal{C}^{inter}$, with $np$ and $ng$ defining role and subgroup counts.

\noindent \paragraph{\textbf{Functional Specifications (FS)}} describe the agents' goals, represented as $\mathcal{FS} = \langle \mathcal{SCH}, \mathcal{PO} \rangle$. The social scheme $\mathcal{SCH}$ includes global goals $\mathcal{G}$, missions $\mathcal{M}$, and plans $\mathcal{P}$ that organize goals in a tree structure. Plans link goals with an operator ($op$) indicating sequence, choice, or parallel completion. Missions map to goal sets ($mo$), and agent counts per mission are specified by $nm$. Preferences $\mathcal{PO}$ indicate which missions agents prefer, denoted as $m_1 \prec m_2$.

\noindent \paragraph{\textbf{Deontic Specifications (DS)}} indicate the relationship between roles goals, given by $\mathcal{DS} = \langle \mathcal{OBL}, \mathcal{PER} \rangle$. Time constraints $\mathcal{TC}$ set periods for permissions or obligations ($Any$ for any time). Obligations ($\mathcal{OBL}$) require agents in role $\rho_a$ to undertake mission $m$ at times $tc$, while permissions ($\mathcal{PER}$) allow it. The $rds$ function maps roles to their deontic specifications as $\langle tc, y, m \rangle$ where $y$ distinguishes permission (0) from obligation (1).

\

\noindent Organizational specifications applied to agents are roles and goals (as missions) through permissions or obligations. Indeed, the other structural specifications such as compatibilities or links are inherent to roles. Similarly, we consider that the goals, the missions, and their mapping ($mo$) are enough to also link all of the other functional specifications such as plans, cardinalities, or preference orders.
Consequently, we consider it is sufficient to take into account roles, missions (goal and mapping) and permissions/obligations when linking $\mathcal{M}OISE^+$ with Dec-POMDP. 

\begin{figure*}[t]
    \label{eq:single_value_function}
    \raggedright
    \textbf{\textit{Definition 1} \quad Sate-Value function adapted to constraint guides in AEC mode:}
    \begin{gather*}
      \text{\quad \quad} V^{\pi^j}(s_t) = \hspace{-0.75cm} \sum_{\textcolor{red}{ \substack{a_{t} \in A \text{ if } rn() < ch_{t}, \\ 
      a_{t} \in A_{t} \text{ else}}
      }}{\hspace{-0.7cm} \pi_i(a_{t} | \omega_t)} \sum_{s_{t+1} \in S}{\hspace{-0.1cm} T(s_{t+1} | s_t, a_{t})[R(s_t,a_{t},s_{t+1}) + \hspace{-0.1cm} \textcolor{blue}{ \sum_{m \in \mathcal{M}_i}{ \hspace{-0.1cm} v_m(t) \frac{grg_m(h_{t+1})}{1 - p + \epsilon} } } + \textcolor{red}{(1-ch_t) \times rrg(\omega_t,a_{t+1})} + V^{\pi^j_{i+1 \ mod \ n}}(s_{t+1})]}
    \end{gather*}  
    \textcolor{red}{\[\text{With } rag(h_t, \omega_t) = A_{t} \times \mathbb{R} \text{, } \langle a_t, ch_{t} \rangle \in A_{t} \times \mathbb{R} \text{ ; } \text{ and } rn: \emptyset \to [0,1[ \text{, a uniform random function}\]}
    \vspace{-0.5cm}
    \textcolor{blue}{
    \begin{gather*}
    \text{With } \omega_t = O(\omega_t | s_t, a_t) \text{ ; } h_t = \{h_0 = \langle \rangle, h_{t+1} = \langle h_t, \langle \omega_{t+1}, a_{t+1} \rangle \rangle \} \text{ ; } grg_m(h) = \hspace{-0.8cm} \sum_{(grg_i,w_i) \in mo(m)}{\hspace{-0.8cm} w_i \times grg_i(h)} \text{ ; } \epsilon \in \mathbb{R}_{>0} \text{ ; }
    \end{gather*}
    }
    \vspace{-0.75cm}
    \textcolor{blue}{
    \begin{gather*}
    v_m(t) = \{ 1 \text{ if } t \in t_c \text{ ; else } 0 \} \text{ ; } \text{ and } \mathcal{M}_i = \{m_j | \langle ar(i),m_j,t_c,p \rangle \in \mathcal{M}\}
    \end{gather*}
    }
    \vspace{-0.6cm}
\end{figure*}

\subsection{Linking $\mathcal{M}OISE^+$ with MARL}

\begin{figure}[h!]
    \centering
    \input{figures/mm_synthesis_single_column.tex}
    \caption{A minimal view of the MOISE+MARL framework: 
    Users first define $\mathcal{M}OISE^+$ specifications, which include roles ($\mathcal{R}$) and missions ($\mathcal{M}$), both associated through $rds$. They then create MOISE+MARL specifications by first defining \textbf{Constraint guides} such as $rag$ and $rrg$ to specify role logic, and $grg$ for goal logic. 
    Next, \textbf{Linkers} are used to connect agents with roles through $ar$ and to link the logic of the constraint guides to the defined $\mathcal{M}OISE^+$ specifications. Once this is set up, roles can be assigned to agents, and the MARL framework updates accordingly during training.
    }
    \label{fig:mm_synthesis}
\end{figure}
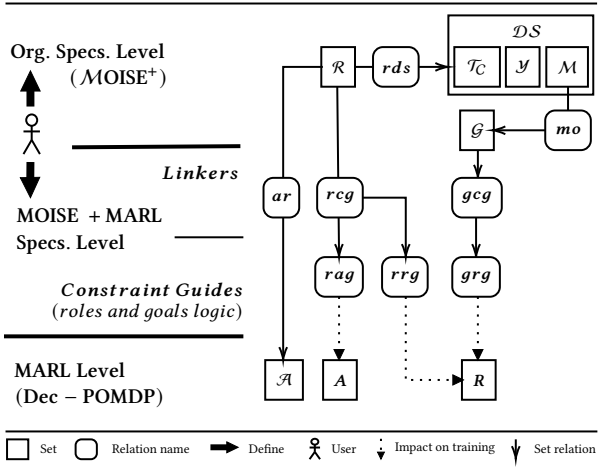

We identified the \textit{AGR}~\cite{ferber2003} (Agent Group Role) and the $\mathcal{M}OISE^+$~\cite{Hubner2007} organizational models. Unlike AGR which is an informal framework introducing roles according to groups, $\mathcal{M}OISE^+$ provides a more detailed and flexible description of the structures and functions of a MAS, easing a formal description of agents' policies in MARL.

\

\noindent The \textbf{Constraint Guides} are three new relations introduced to describe the logic of the roles and goals of $\mathcal{M}OISE^+$ in the Dec-POMDP formalism:
%
% \begin{itemize}
\begin{enumerate*}[label={\roman*) },itemjoin={; \quad}]

    \item \textbf{Role Action Guide} \quad $rag: H \times \Omega \rightarrow \mathcal{P}(A \times \mathbb{R})$, the relation that models a role as a set of rules which, for each pair consisting of a history $h \in H$ and an observation received by the agent $\omega \in \Omega$, associates expected actions $A \in \mathcal{P}(A)$ each associated with a constraint hardness $ch \in [0,1]$ ($ch = 1$ by default). By restricting the choice of the next action among those authorized, the agent is forced to adhere to the expected behavior of the role
    \item \textbf{Role Reward Guide} \quad $rrg: H \times \Omega \times A \to \mathbb{R} = \{r_m \text{ if } a \notin A_\omega \text{, } rag(h, \omega) \allowbreak = \allowbreak A_\omega \times \mathbb{R} \text{, } h \in H; \text{ else } 0\}$, the relation that models a role by adding a penalty $r_m$ to the global reward if the last action chosen by the agent $a \in A$ is not authorized. This is intended to encourage the agent to adhere to the expected behavior of a role
    \item \textbf{Goal Reward Guide} \quad $grg: H \rightarrow \mathbb{R}$, the relation that models a goal as a soft constraint by adding a bonus $r_b \in \mathbb{R}$ to the global reward if the agent's history $h \in H$ contains a characteristic sub-sequence $h_g \in H_g$ of the goal, encouraging the agent to reach it.
\end{enumerate*}
% \end{itemize}

\

\noindent Finally, we introduce the \textbf{Linkers} to link the $\mathcal{M}OISE^+$ organizational specifications with constraint guides and agents:
%
% \begin{itemize}
\begin{enumerate*}[label={\roman*) },itemjoin={; \quad}]

    \item \textbf{Agent to Role} \quad $ar: \mathcal{A} \to \mathcal{R}$, the bijective relation linking an agent to a role;
    \item \textbf{Role to Constraint Guide} \quad $rcg: \mathcal{R} \rightarrow rag \cup rrg$, the relation associating each $\mathcal{M}OISE^+$ role to a $rag$ or $rrg$ relation, forcing/encouraging the agent to follow the expected actions for the role $\rho \in \mathcal{R}$;
    \item \textbf{Goal to Constraint Guide} \quad $gcg: \mathcal{G} \rightarrow grg$, the relation linking goals to $grg$ relations, representing goals as rewards in MARL.
\end{enumerate*}
% \end{itemize}

\paragraph{\textbf{Resolving the MOISE+MARL problem}}
% formalized as $MM = \langle D, \mathcal{OS}\allowbreak, ar, rcg, \allowbreak gcg, rag, rrg, grg\rangle$
involves finding a joint policy $\pi^{j} = \{\pi^j_0,\pi^j_1\dots\pi^j_n\}$ that maximizes the state-value function $V^{\pi^{j}}$ (or reaches a minimum threshold), which represents the expected cumulative reward starting from an initial state $s \in S$ and following the joint policy $\pi^{j}$, applying successive joint actions $a^{j} \in A^n$ under additional constraint guides. The state-value is described in the case where agents act sequentially and cyclically (Agent Environment Cycle - AEC mode) in \hyperref[eq:single_value_function]{Definition 1}, adapting its definition for roles (in red) and missions (in blue), impacting the action space and reward. \autoref{fig:mm_synthesis} illustrates the links between $\mathcal{M}OISE^+$ and Dec-POMDP via the MOISE+MARL framework.

At any time $t \in \mathbb{N}$ (initially $t = 0$), the agent $i = t \ mod \ n$ is constrained to a role $\rho_i = ar(i)$. For each temporally valid deontic specification $d_i = rds(\rho_i) = \langle tc_i,y_i, m_i \rangle$, the agent is permitted (if $y_i = 0$) or obligated (if $y_i = 1$) to commit in mission $m_i \in \mathcal{M}, \mathcal{G}_{m_i} = mo(m_i)$, and $n \in \mathbb{N}$ the number of agents.
First, based on the received observation $\omega_t$, the agent must choose an action either: within the expected actions of the role $A_t$ if a random value is below the role constraint hardness $ch_t$; or within the set of all actions $A$ otherwise. If $ch_t = 1$, the role is strongly constrained for the agent and weakly otherwise.
Then, the action is applied to the current state $s_t$ to transition to the next state $s_{t+1}$, generate the next observation $\omega_{t+1}$, and yield a reward. The reward is the sum of the global reward with penalties and bonuses obtained from the organizational specifications: \quad i) the sum of the bonuses for goals associated with each temporally valid mission (via Goal Reward Guides), weighted by the associated value ($\frac{1}{1-p+\epsilon}$); \quad ii) the penalty associated with the role (via "Role Reward Guides") weighted by the role constraint hardness.
Finally, the cumulative reward calculation continues in the next state $s_{t+1} \in S$ with the next agent $(i+1) \ mod \ n$.

\subsection{Easying constraint guides implementation}

Since roles, goals, and missions as simple labels, their definition is assumed. However, implementing a $rag$, $rrg$, or $grg$ relation requires defining a potentially large number of histories, possibly redundant. Therefore, an extensional definition of a set of histories can be tedious. Moreover, the logic of all constraint guides takes the agent trajectory as input to determine whether the trajectory belongs to a predefined history set. For example, a $rag$ relation can be seen as determining the next expected actions depending on whether the trajectory belongs to a given set and the new observation received.

A first approach is to let users develop their constraint guides in an intensional way with custom logic (such as a script code) in order to analyse history and compute the output in a manageable way. In that case, the relation $b_g: H \to \{0,1\}$ formalizes how users propose to determine whether a history belongs to a predefined set $H_g$.
To help implement this relation, we propose a \textbf{Trajectory-based Pattern} (TP) inspired by Natural Language Processing, denoted $p \in P$, as a way to define a set of histories in an intensional way.

A TP implies that any considered real observation or action is known and mapped to a label $l \in L$ (through $l: \Omega \cup A \to L$) to be conveniently managed. A TP $p \in P$ is defined as follows: $p$ is: either a "leaf sequence" denoted as a couple of history-cardinality $s_l = \langle h, \{c_min,c_max\}\rangle$ (where $h \in H, c_{min} \in \mathbb{N}, c_{max} \in \mathbb{N} \cup "*")$; or a "node sequence" denoted as a couple of a tuple of concrete sequences and cardinality $s_n = \langle \langle s_{l_1}, s_{l_1}\dots \rangle, \{c_min,c_max\}\rangle$. For example, the pattern $p = \allowbreak "[o_1,a_1,[o_2,a_2]\langle0,2\rangle]\langle1,*\rangle"$ can be formalized as the node sequence $\allowbreak \langle \langle \langle o_1,a_1\rangle,\langle 1,1 \rangle \rangle, \langle \langle o_2,a_2 \rangle, \langle 0,2 \rangle \rangle \rangle \langle 1,"*" \rangle$, indicating the set of histories $H_p$ containing at least once the sub-sequence consisting of a first pair $\langle o_1,a_1\rangle$ and then at most two repetitions of the pair $\langle o_2,a_2 \rangle$.
The relation $b_g$ then becomes $b_g(h) = m(p_g,h), \text{ with } m: P \times H \to \{0,1\}$ indicating if a history $h \in H$ matches a history pattern $p \in P$ describing a history set $H_g$.

\section{The TEMM method}
\label{sec:TEMM_algorithm}

As presented in \autoref{sec:related_works}, we were unable to identify any available method that fully meets our requirements for determining implicit roles, implicit goals, or organizational fit. Therefore, we propose the \textbf{Trajectory-based Evaluation in MOISE+MARL} (TEMM) method for automatic inference and evaluation of roles and missions.
TEMM uses unsupervised learning techniques to generalize roles and missions from the set of collected trajectories over multiple test episodes. By measuring the gap between inferred implicit organizational specifications and actual behaviors, we can also quantify the organizational fit as to how well a policy conforms to the inferred implicit organizational specifications.

TEMM is based on proposed definitions for each $\mathcal{M}OISE^+$ organizational specification regarding joint-histories or other organizational specifications, using specific unsupervised lea-rning techniques to infer them progressively. Here, we provide an informal description of the method~\hyperref[fn:github]{\footnotemark[1]}.
\footnotetext[1]{ \label{fn:github} Additional details, developed code, datasets containing all the hyperparameters and details of the organizational specifications are available at \url{https://github.com/julien6/MOISE-MARL}}

\paragraph{\textbf{1) Inferring roles and their inheritance}}

We introduce that a role $\rho$ is defined as a policy whose associated agents' histories all contain a Common Longest Sequence (CLS). We introduce that a role $\rho_2$ inherits from $\rho_1$ if the CLS of histories associated with $\rho_2$ is also contained within that of $\rho_1$.
Based on these definitions, TEMM uses a "hierarchical clustering" technique to find the CLSs among agent histories. The results can be represented as a dendrogram, allowing inferring implicit roles and inheritance relationships, their respective relationships with histories.
We measure the gap between current agents' sequence and inferred implicit roles' sequences, as the "structural organizational fit".

\paragraph{\textbf{2) Inferring goals, plans, and missions}}

We introduce that a goal is a set of common joint-observation reached by following the histories of successful agents.
For each joint-history, TEMM calculates the joint-observation transition graph, which is then merged into a general graph. By measuring the distance between two vectorized joint-observations with K-means, we can find trajectory clusters that some agents may follow. Then, we sample some sets of joint-observations for each trajectory as implicit goals. For example, we can select the narrowest set of joint-observations where agents seem to collectively transition at a given time to reach their goal. Otherwise, balanced sampling on low-variance trajectories could be performed. Knowing which trajectory a goal belongs to, TEMM infers plans based solely on choices and sequences.

We introduce that a mission is the set of goals that one or more agents are accomplishing.
Knowing the shared goals achieved by the agents, TEMM determines representative goal sets as missions.
By measuring the distance between inferred implicit goals which joint-observations with current agents' joint-observation, we compute the "structural organizational fit".

\paragraph{\textbf{3) Inferring obligations and permissions}}

We introduce that an obligation is when an agent playing the role $\rho$ fulfills the goals of a mission and no others during certain time constraints, while permission is when the agent playing the role $\rho$ may fulfill other goals during specific time constraints.
TEMM determines which agents are associated with which mission and whether they are restricted to certain missions, making them obligations, or if they have permission.
Having already computed structural organizational fit and functional organizational fit, the organizational fit is the sum of these two values.

\

Overall, the K-mean and hierarchical clustering techniques require manual configuration to obtain roles and goals, avoiding introducing perturbations that could lead to determining false organizational specifications. Despite this, the method recommends thoroughly understanding the obtained roles and goals to manually identify and remove any remaining perturbations.

\section{Experimental framework}
\label{sec:experimental_setup}

This section details the experimental framework used to evaluate the MOISE+MARL framework.% We adapted existing tools to implement our approach. We then present the environments used, the MARL algorithms, the organizational specifications, and the evaluation metrics.

\subsection{Implementing MOISE+MARL}

We have developed an implementation of the MOISE+MARL framework called \textquote{MMA}~\hyperref[fn:github]{\footnotemark[1]} (MOISE+MARL API), which is a Python API that integrates all theoretical sets and relations to minimize user interactions. MMA uses an Object-oriented approach, structuring the $\mathcal{M}OISE^+$ model as nested data classes, with the "Moise" class at the root, enabling users to define organizational specifications, such as roles, goals, and permissions.

To support Dec-POMDP environments, we utilized the \textit{PettingZoo} library \cite{terry2020pettingzoo}, which provides a standard API for multi-agent systems and ensures interoperability across various environments, similar to the Gymnasium framework \cite{kwiatkowski2024}. MMA incorporates a dictionary for observation/action label mapping ($l$), which users can customize, and it also supports Trajectory Patterns (TPs) to facilitate pattern definition and matching.

Each type of constraint guide, like $rag$, $rrg$, and $grg$, is implemented as a separate class. Users can define these guides with custom functions or JSON rules; for example, $rag$ can be instantiated by associating a $\langle \text{TP, last observation} \rangle$ pair with expected actions, while $grg$ can apply bonuses based on specific TPs. The global "MMA" class integrates these guides with user-defined relations, such as linking an agent to a role ($ar$) or associating a role with $rrg$ and $rag$, incorporating the organizational specifications defined in the $\mathcal{M}OISE^+$ structure.

Once set up, the MMA object is used to encapsulate the environment with a \textit{PettingZoo} wrapper. This wrapper applies action masks and modifies rewards at each step, ensuring that agents adhere to the organizational specifications throughout training. MMA also integrates \textit{MARLlib} \cite{hu2021marlib}, which provides access to state-of-the-art MARL algorithms, enabling training to be run on a high-performance computing cluster.

After training, the TEMM method is employed, using manually optimized hyperparameters to infer implicit roles and goals through hierarchical clustering and K-means. This analysis generates visual outputs, such as dendrograms for roles and joint-observation transition graphs for goals. The resulting implicit roles and goals can be exported as JSON trajectories, providing a structured view of the inferred organizational behaviors.

\subsection{Environments used}

We test MOISE+MARL in four different MARL environments, each modeled as a Dec-POMDP simulation scenario. These environments were selected for their diversity in terms of collaboration and resource management. Here is a description of each:

\begin{itemize}
% \begin{enumerate*}[label={\roman*) },itemjoin={; \quad}]

    \item \textbf{Predator-Prey}: A classic environment where several predators must cooperate to capture prey. This environment tests the agents' ability to coordinate their actions to achieve a collective goal\cite{lowe2017multi}

    \item \textbf{Overcooked-AI}: A team cooking game where several agents must collaborate to prepare and serve dishes in increasingly complex kitchens\cite{overcookedai}. Agents must manage tasks such as chopping, cooking, assembling, and serving ingredients while optimizing their movements and avoiding obstacles. This environment is ideal for testing coordination and task allocation in dynamic, highly interdependent scenarios, where clear roles (such as "chef," "assistant," "server") can be defined via organizational specifications
    
    \item \textbf{Warehouse Management}: A proposed environment, where agents must manage a warehouse by coordinating resource deliveries to demand points. Roles and missions here influence agent specialization in specific tasks (transportation of products, inventory management)
    
    \item \textbf{Cyber-Defense Simulation}: A complex environment si-mulating network defense against cyberattacks. Agents must identify and counter threats while adhering to strict security rules, thus testing the safety of trained agents\cite{Maxwell2021}.
% \end{enumerate*}
\end{itemize}

These environments are encapsulable in the PettingZoo API, enabling seamless integration with our MOISE+MARL implementation and facilitating the application of organizational specifications.

\subsection{MARL algorithms used}

We evaluated our framework with several MARL algorithms :
\begin{enumerate*}[label={\roman*) },itemjoin={; \quad}]

    \item \textbf{MADDPG (Multi-Agent Deep Deterministic Policy Gradient)}~\cite{lowe2017multi}: A centralized learning, decentralized execution algorithm, allowing each agent to have a deterministic policy while using global information during training
    
    \item \textbf{MAPPO (Multi-Agent Proximal Policy Optimization)} \cite{yu2021mappo}: An adapted version of PPO for MAS, optimized for stable joint policy convergence in complex scenarios
    
    \item \textbf{Q-Mix}~\cite{rashid2018qmix}: A Q-value-based algorithm that learns to combine individual agents' Q-values into a joint value to optimize cooperation
    
    \item \textbf{COMA (Counterfactual Multi-Agent) }~\cite{foerster2018counterfactual} An actor-critic algorithm able to estimate the impact of an individual agent's actions on the team's overall reward.
\end{enumerate*}

\subsection{Organizational specifications}

For each environment, we defined a set of organizational specifications. These specifications include roles, missions, as well as permissions and obligations. Here, we give an informal description of these~\hyperref[fn:github]{\footnotemark[1]}:
%
% \begin{itemize}
\begin{enumerate*}[label={\roman*) },itemjoin={; \quad}]

    \item \textbf{Predator-Prey}: Predator and prey roles are defined, with each predator having specific goals such as "capture the prey" or "block escape routes."

    \item \textbf{Overcooked-AI}: Agents adopt three main roles: chef, assistant, and server. The Chef is responsible for cooking and assembling dishes, the Assistant handles ingredient chopping and supply, and the Server is in charge of delivering dishes to customers. Missions primarily involve preparing and serving a certain number of dishes within a given time.
    
    \item \textbf{Warehouse Management}: Agents adopt roles such as "transporter" and "inventory manager," with missions related to managing logistics flows and optimized delivery.
    
    \item \textbf{Cyber-Defense Simulation}: Agents have network defender roles, each with obligations such as intrusion detection or protecting specific drone swarm ad hoc networks.
\end{enumerate*}
% \end{itemize}

\subsection{Computing resources and hyperparameters}

All experiments were conducted on an academic high-performance computing cluster, utilizing various configurations of GPU nodes. Specifically, we employed nodes equipped with NVIDIA A100 and V100 GPUs, and AMD MI210 GPUs. Each algorithm-environment combination was executed on 5 parallel instances to ensure robust and consistent results.
Hyperparameters~\hyperref[fn:github]{\footnotemark[1]} for each algorithm, including learning rates, discount factors, and exploration rates, were either retrieved from MARLlib data banks or optimized for each environment through a grid search using the \textit{Optuna} tool~\cite{akiba2019optuna}.

\subsection{Evaluation metrics and protocol}

To measure the policy effectiveness and the impact of organizational specifications, we defined the following metrics:
\begin{enumerate*}[label={\roman*)}, itemjoin={; \quad}]
% \begin{itemize}
    \item \textbf{Cumulative Reward}: Measures policy effectiveness in achieving environment goals
    \item \textbf{Reward Standard Deviation}: Reflects the stability of learned policies over episodes
    \item \textbf{Convergence Rate}: Indicates the speed at which policies achieve stable performance
    \item \textbf{Constraint Violation Rate}: Assesses policy adherence to organizational constraints, critical for safety
    \item \textbf{Consistency Score}: Evaluates alignment between trained behaviors and organizational specifications
    \item \textbf{Robustness Score}: Measures agents' ability to maintain performance under a series of challenging scenarios
    \item \textbf{Organizational Fit Level}: Quantifies the organizational fit.
% \end{itemize}
\end{enumerate*}

\

\noindent Our protocol compares the \textit{Reference Baseline} (RB) without organizational constraints and the \textit{Organizationally Constrained Baseline} (OB) using MOISE+MARL.

We use the MMA software to establish the RB with no organizational specifications. For each environment, we train agents with each algorithm until rewards converge or a maximum episode limit is reached. We record metrics and select the algorithm that achieves the highest Cumulative Reward as the RB (control scenario without constraints).
For the OB, we reset environments and agents, applying pre-defined organizational specifications using MMA so that each agent is assigned a role. We train these agents with the RB's highest-performing algorithm, again until convergence or the episode limit. After training, we compute all metrics, providing a scenario with organizational constraints as the OB.

By comparing the RB and OB, we can validate the impact of MOISE+MARL on organizational fit. First, we check if the agents' behaviors align with the specified roles in the OB. We analyze manually or rely on reliable metrics like Reward Standard Deviation, Convergence Rate, and Robustness Score. If agents behave in ways that align with their roles, then we favor the idea that MOISE+MARL has influenced organizational fit.
Therefore, we should observe differences in the Organizational Fit Level metric between RB and OB. We can also push forward a correlation between fully/freely constraining roles and higher/lower Organizational Fit Level. If all of these observations hold, then the Organizational Fit Level may quantify the organizational fit, and the Consistency Score metric may be used to validate the effectiveness of MOISE+MARL in controlling organizational fit when roles are applied.

Finally, we also check the relevance of the $\mathcal{M}OISE^+$ by comparing MOISE+MARL with its AGR equivalent called AGR+MARL which only considers roles and  does not explicitly include goals.

\section{Results}
\label{sec:results}

This section presents and analyzes the experimental results from applying MOISE+MARL across the environments.%, highlighting key metrics and comparisons with AGR+MARL.

\begin{table*}[h!]
    \centering
    \caption{Detailed results for each environment and favored algorithm under both RB and OB.}
    \label{tab:detailed_results}
    \small
    \renewcommand{\arraystretch}{1.2}
    \begin{tabular}{p{3.5cm}p{1.5cm}p{1.cm}p{1.3cm}p{1cm}p{1.3cm}p{1.3cm}p{1.2cm}p{1.2cm}p{1cm}}
        \hline
        \textbf{Env.} & \textbf{Alg.} & \textbf{Org. Spec.} & \textbf{Cum. Rew.} & \textbf{STD} & \textbf{Conv. Rate} & \textbf{Viol. Rate} & \textbf{Cons. Score} & \textbf{Rob. Score} & \textbf{Org. Fit Lvl} \\ \hline
        Predator-Prey & MADDPG &  & 200.1 & 21.5 & 0.65 & 12.3\% & - & 0.65 & 0.43 \\
        Predator-Prey & MADDPG & Yes & 245.8 & 15.2 & 0.85 & .0\% & 0.81 & 0.83 & 0.87 \\
        Overcooked-AI & MAPPO &  & 348.2 & 15.6 & 0.75 & 7.1\% & - & 0.71 & 0.48 \\
        Overcooked-AI & MAPPO & Yes & 391.2 & 10.4 & 0.92 & .0\% & 0.89 & 0.89 & 0.91 \\
        Warehouse Management & Q-Mix &  & 257.4 & 18.9 & 0.74 & 7.8\% & - & 0.68 & 0.50 \\
        Warehouse Management & Q-Mix & Yes & 307.1 & 13.8 & 0.88 & .0\% & 0.88 & 0.86 & 0.90 \\
        Cyber-Defense & COMA &  & 162.4 & 17.3 & 0.70 & 12.2\% & - & 0.67 & 0.45 \\
        Cyber-Defense & COMA & Yes & 188.9 & 11.2 & 0.86 & .0\% & 0.76 & 0.80 & 0.83 \\ \hline
    \end{tabular}
\end{table*}

\subsection{Quantitative organizational fit and consistency}

\autoref{tab:detailed_results} summarizes the performance metrics for each environment and the most efficient algorithm under both the RB and OB. Across all environments, the organizational fit metric is significantly higher under the OB, confirming that MOISE+MARL effectively aligns agent behaviors with organizational specifications.

For example, in the \textbf{Predator-Prey} environment with \textbf{MADDPG}, agents in the OB configuration achieved an organizational fit level of 0.87, which represents a 44\% increase compared to the RB (0.43). Similarly, in the \textbf{Overcooked-AI} environment, \textbf{MAPPO} under the OB reached an organizational fit of 0.91 (an increase of 89\% over the RB's 0.48). These improvements are mirrored in the \textbf{Warehouse Management} environment with \textbf{Q-Mix}, where the organizational fit rose from 0.50 in the RB to 0.90 in the OB, suggesting a MOISE+MARL's consistent effectiveness.

In general, agents constrained with organizational specifications show a lower reward deviation and a higher convergence rate that suggests an impact on their behavior. We manually observed agents' interactions in visualizable environments such as Predator-Prey and verified that trained agents' behaviors do align with the expected behavior of a structural and functional implicit organization.
Indeed, the significant variation depending on the application of organizational specifications on agents, and the manually verified alignment of agents with roles suggests that organisational fit level correlates with the organizational fit.

Considering organizational fit level reliable across all environments, the \textbf{consistency score} also shows important values with a minimal value of 0.76 for the \textbf{Cyber-Defense} environment. This suggests that despite a noisy environment that introduces some disturbance in agents' behavior, the inferred organizational specifications are still close to applied ones.

\subsection{Performance and stability across algorithms}

The results indicate that policy-based and actor-critic algorithms like \textbf{MADDPG} and \textbf{MAPPO} benefit substantially from the MOISE+ MARL framework, particularly in terms of consistency and stability. For example, \textbf{MAPPO} in the \textbf{Overcooked-AI} environment saw a reward standard deviation reduction from 15.6 (RB) to 10.4 (OB), reflecting a more stable policy with less behavioral fluctuation. \textbf{MADDPG} in \textbf{Predator-Prey} also showed a similar pattern, with a standard deviation drop from 21.5 in the RB to 15.2 in the OB, indicating increased reliability.

In contrast, value-based algorithms like \textbf{Q-Mix} maintained high performance in cumulative reward but displayed greater variability in consistency. For instance, in \textbf{Warehouse Management}, \textbf{Q-Mix} achieved a reward standard deviation of 13.8 in the OB, a notable improvement over 18.9 in the RB but still higher than the stability observed in policy-based algorithms. This suggests that while \textbf{Q-Mix} is effective for achieving task goals, it may require further tuning for roles with MOISE+MARL to enhance consistency.

\subsection{Impact of organizational constraints on policy convergence, robustness and violation rates}

Applying organizational constraints resulted in faster convergence rates across all environments. In the \textbf{Cyber-Defense} environment, \textbf{COMA} with MOISE+MARL converged at a rate of 0.86, compared to 0.70 in the RB. Similar trends were observed in the \textbf{Warehouse Management} environment with \textbf{Q-Mix}, which showed an improvement from 0.74 in the RB to 0.88 in the OB. This expedited convergence can be attributed to the structured guidance of roles and missions, which narrows the policy search space.

In addition to the presented results where constraint hardness is set to 1, we observed that constraint violation rates were consistently higher when organizational constraints were defined with a lower constraint hardness. In \textbf{Overcooked-AI}, \textbf{MAPPO} recorded a null violation rate with a constraint hardness of 1, compared to 7.1\% with a constraint hardness of 0. Similarly, in \textbf{Warehouse Management}, \textbf{Q-Mix} reduced the violation rate from 7.8\% to zero as constraint hardness increased. This further supports the framework's effectiveness in enhancing adherence to desired behaviors.

Additionally, we observed a consistent improvement in robustness when organizational specifications were applied to agents. For instance, \textbf{MADDPG} in \textbf{Predator-Prey} and \textbf{MAPPO} in \textbf{Overcooked-AI} achieved high consistency scores of 0.81 and 0.89, respectively, indicating that agents closely followed the inferred roles. Robustness also improved, with \textbf{MAPPO} in \textbf{Overcooked-AI} achieving a robustness score of 0.89, up from 0.71 in the RB, underscoring the framework's impact on agents' resilience to perturbations.

However, one can point out a potential bias: organizational specifications were specifically designed to encompass all observations, avoiding non-handled new situations.

\subsection{Comparison between MOISE+MARL and AGR+MARL}

\begin{table}[h!]
    \centering
    \caption{Performance comparison between MOISE+MARL and AGR+MARL.}
    \label{tab:ablation_study}
    \small
    \renewcommand{\arraystretch}{1.1}
    \begin{tabular}{p{2cm}p{0.5cm}p{0.6cm}p{1.3cm}p{0.6cm}p{1.3cm}}
        \hline
        \textbf{Framework} & \textbf{Env.} & \textbf{Conv. Rate} & \textbf{Robustness Score} & \textbf{Org. Fit} & \textbf{Cumulative Reward} \\ \hline
        MOISE+MARL & PP & 0.85 & 0.83 & 0.87 & 245.8 \\
        AGR+MARL & PP & 0.75 & 0.69 & 0.56 & 208.4 \\
        MOISE+MARL & OA & 0.92 & 0.89 & 0.91 & 391.2 \\
        AGR+MARL & OA & 0.82 & 0.75 & 0.58 & 348.9 \\
        MOISE+MARL & WM & 0.88 & 0.86 & 0.90 & 307.1 \\
        AGR+MARL & WM & 0.76 & 0.72 & 0.61 & 278.6 \\ \hline
    \end{tabular}
\end{table}

\noindent \autoref{tab:ablation_study} highlights the impact of intermediary goals within MOISE+ MARL. In \textbf{Overcooked-AI}, \textbf{MAPPO} under MOISE+MARL achieved a cumulative reward of 391.2, with an organizational fit of 0.91—33\% higher than AGR+MARL's 0.58. Similarly, in \textbf{Warehouse Management}, \textbf{Q-Mix} under MOISE+MARL attained a cumulative reward of 307.1, an increase of nearly 10\% over AGR+MARL's 278.6, with a higher robustness score (0.86 vs. 0.72).

Overall, these results underscore the importance of intermediary goals in fostering more stable, goal-oriented behaviors. By facilitating a clearer path to the global goal, MOISE+MARL consistently outperforms AGR+MARL in achieving higher rewards, robustness, and organizational fit across Predator-Prey (PP), Warehouse Management (WM), and Overcooked-AI (OA).
Finally, we analyzed the impact of increasing the number of organizational constraints on training time. Preliminary results suggest a nearly linear growth in training duration as the number of constraints increases~\footnotemark[1].

\section{Conclusion and future works}
\label{sec:discussion_conclusion_future_work}

The MOISE+MARL framework introduced in this paper aims to enhance control and explainability in MARL by incorporating organizational models that define explicit roles and missions for agents. Experimental results across several environments indicate that this framework helps agents adhere to expected behaviors while facilitating better policy convergence by constraining the policy search space. The results also show that agents trained with roles and goals exhibit behaviors closely resembling those determined via the framework, suggesting coherence between the application of organizational specifications and their expected effects.

However, the framework's reliance on predefined organizational specifications means it may struggle to adapt in highly dynamic or unstructured environments where agent roles and missions are less defined or evolve over time.
Moreover, the computational overhead associated with enforcing organizational constraints and dynamically modifying rewards and actions may pose scalability challenges. Additionally, TEMM can be computationally intensive, which may hinder its applicability in real-time scenarios.

We are currently pursuing three main directions:
%
% \begin{enumerate*}[label={\roman*)}, itemjoin={; \quad}]
\begin{itemize}
    \item Developing adaptive mechanisms that allow roles and missions to evolve dynamically during training, enabling agents to respond to changes in real-time
    \item Exploring automated methods, such as Large Language Models, for generating organizational specifications based on observed agent behaviors to help users on defining these specifications manually
    \item Improving the computational efficiency of TEMM or exploring alternative evaluation methods for real-world applications with larger agent populations.
\end{itemize}
% \end{enumerate*}

\begin{acks}
  This work was supported by \emph{Thales Land Air Systems} within the framework of the \emph{Cyb'Air} chair and the \emph{AICA IWG}
\end{acks}

%%%%%%%%%%%%%%%%%%%%%%%%%%%%%%%%%%%%%%%%%%%%%%%%%%%%%%%%%%%%%%%%%%%%%%%%

%%% The next two lines define, first, the bibliography style to be 
%%% applied, and, second, the bibliography file to be used.

\bibliographystyle{ACM-Reference-Format}
\balance
\bibliography{references}

%%%%%%%%%%%%%%%%%%%%%%%%%%%%%%%%%%%%%%%%%%%%%%%%%%%%%%%%%%%%%%%%%%%%%%%%

\end{document}

%% file: figures/moise_model.tex
\tikzset{every picture/.style={line width=0.75pt}} %set default line width to 0.75pt        

\begin{tikzpicture}[x=0.75pt,y=0.75pt,yscale=-1,xscale=1]
%uncomment if require: \path (0,2639); %set diagram left start at 0, and has height of 2639

%Shape: Rectangle [id:dp6515425466042692] 
\draw  [fill={rgb, 255:red, 255; green, 255; blue, 255 }  ,fill opacity=1 ] (176,834) -- (462,834) -- (462,957) -- (176,957) -- cycle ;
%Shape: Rectangle [id:dp5082212387639646] 
\draw  [fill={rgb, 255:red, 184; green, 233; blue, 134 }  ,fill opacity=0.34 ] (362,842) -- (457.1,842) -- (457.1,952) -- (362,952) -- cycle ;
%Shape: Rectangle [id:dp5297773644830988] 
\draw  [fill={rgb, 255:red, 184; green, 233; blue, 134 }  ,fill opacity=0.34 ] (366,868) -- (453.83,868) -- (453.83,948) -- (366,948) -- cycle ;
%Shape: Rectangle [id:dp1575993513067182] 
\draw  [fill={rgb, 255:red, 248; green, 231; blue, 28 }  ,fill opacity=0.4 ] (179.27,838) -- (298,838) -- (298,952) -- (179.27,952) -- cycle ;
%Straight Lines [id:da8705971767634542] 
\draw    (357.69,930) -- (372,930) ;
\draw [shift={(374,930)}, rotate = 180] [color={rgb, 255:red, 0; green, 0; blue, 0 }  ][line width=0.75]    (4.37,-1.96) .. controls (2.78,-0.92) and (1.32,-0.27) .. (0,0) .. controls (1.32,0.27) and (2.78,0.92) .. (4.37,1.96)   ;
\draw [shift={(355.69,930)}, rotate = 0] [color={rgb, 255:red, 0; green, 0; blue, 0 }  ][line width=0.75]    (4.37,-1.96) .. controls (2.78,-0.92) and (1.32,-0.27) .. (0,0) .. controls (1.32,0.27) and (2.78,0.92) .. (4.37,1.96)   ;
%Shape: Rectangle [id:dp6422856240164194] 
\draw  [fill={rgb, 255:red, 248; green, 231; blue, 28 }  ,fill opacity=0.4 ] (183.81,866) -- (294,866) -- (294,947) -- (183.81,947) -- cycle ;
%Shape: Rectangle [id:dp011814159340156172] 
\draw  [fill={rgb, 255:red, 248; green, 231; blue, 28 }  ,fill opacity=0.4 ] (216.59,919) -- (237.41,919) -- (237.41,939) -- (216.59,939) -- cycle ;

%Shape: Rectangle [id:dp4100462490677821] 
\draw  [fill={rgb, 255:red, 248; green, 231; blue, 28 }  ,fill opacity=0.4 ] (188,919) -- (214,919) -- (214,939) -- (188,939) -- cycle ;
%Shape: Rectangle [id:dp4224844209889693] 
\draw  [fill={rgb, 255:red, 248; green, 231; blue, 28 }  ,fill opacity=0.4 ] (187.02,892) -- (226.03,892) -- (226.03,912) -- (187.02,912) -- cycle ;

%Shape: Rectangle [id:dp2109151755211085] 
\draw  [fill={rgb, 255:red, 248; green, 231; blue, 28 }  ,fill opacity=0.4 ] (186.22,870) -- (225.8,870) -- (225.8,890) -- (186.22,890) -- cycle ;

%Shape: Rectangle [id:dp5612046560441226] 
\draw  [fill={rgb, 255:red, 248; green, 231; blue, 28 }  ,fill opacity=0.4 ] (251.83,870) -- (291.95,870) -- (291.95,890) -- (251.83,890) -- cycle ;

%Shape: Rectangle [id:dp8653779879037786] 
\draw  [fill={rgb, 255:red, 248; green, 231; blue, 28 }  ,fill opacity=0.4 ] (251.29,892) -- (291.72,892) -- (291.72,912) -- (251.29,912) -- cycle ;

%Shape: Rectangle [id:dp5118226471482248] 
\draw  [fill={rgb, 255:red, 65; green, 117; blue, 5 }  ,fill opacity=1 ] (374.01,917) -- (391.99,917) -- (391.99,937) -- (374.01,937) -- cycle ;

%Shape: Rectangle [id:dp9933924068366515] 
\draw  [fill={rgb, 255:red, 248; green, 231; blue, 28 }  ,fill opacity=0.4 ] (242.63,919) -- (263.37,919) -- (263.37,939) -- (242.63,939) -- cycle ;

%Shape: Rectangle [id:dp8500726713846194] 
\draw  [fill={rgb, 255:red, 248; green, 231; blue, 28 }  ,fill opacity=0.4 ] (183.75,843) -- (206.25,843) -- (206.25,863) -- (183.75,863) -- cycle ;

%Shape: Rectangle [id:dp5924732056565294] 
\draw  [fill={rgb, 255:red, 248; green, 231; blue, 28 }  ,fill opacity=0.4 ] (264.98,842) -- (289.16,842) -- (289.16,862) -- (264.98,862) -- cycle ;

%Shape: Rectangle [id:dp346383687870204] 
\draw  [fill={rgb, 255:red, 74; green, 144; blue, 226 }  ,fill opacity=1 ] (309.84,891) -- (348.24,891) -- (348.24,911) -- (309.84,911) -- cycle ;

%Shape: Rectangle [id:dp8336033635496647] 
\draw  [fill={rgb, 255:red, 74; green, 144; blue, 226 }  ,fill opacity=1 ] (310.57,919) -- (349.48,919) -- (349.48,939) -- (310.57,939) -- cycle ;

%Shape: Rectangle [id:dp7738606236066947] 
\draw  [fill={rgb, 255:red, 184; green, 233; blue, 134 }  ,fill opacity=0.34 ] (366.75,845) -- (395.25,845) -- (395.25,865) -- (366.75,865) -- cycle ;

%Shape: Rectangle [id:dp7635846319928088] 
\draw  [fill={rgb, 255:red, 184; green, 233; blue, 134 }  ,fill opacity=0.34 ] (382.43,891) -- (408,891) -- (408,911) -- (382.43,911) -- cycle ;

%Shape: Rectangle [id:dp3026483542328955] 
\draw  [fill={rgb, 255:red, 184; green, 233; blue, 134 }  ,fill opacity=0.34 ] (419.11,892) -- (444.29,892) -- (444.29,912) -- (419.11,912) -- cycle ;

%Shape: Rectangle [id:dp8981167883687693] 
\draw  [fill={rgb, 255:red, 184; green, 233; blue, 134 }  ,fill opacity=0.34 ] (432.02,917) -- (450,917) -- (450,937) -- (432.02,937) -- cycle ;

%Shape: Rectangle [id:dp9543580464730599] 
\draw  [fill={rgb, 255:red, 184; green, 233; blue, 134 }  ,fill opacity=0.34 ] (404,917) -- (421.98,917) -- (421.98,937) -- (404,937) -- cycle ;

%Shape: Rectangle [id:dp9228172603845481] 
\draw  [fill={rgb, 255:red, 245; green, 166; blue, 35 }  ,fill opacity=1 ] (266,919) -- (286.77,919) -- (286.77,939) -- (266,939) -- cycle ;

%Shape: Rectangle [id:dp22226375116176755] 
\draw  [fill={rgb, 255:red, 255; green, 255; blue, 255 }  ,fill opacity=1 ] (261.49,962) -- (271.29,962) -- (271.29,974) -- (261.49,974) -- cycle ;
%Straight Lines [id:da8264214166350665] 
\draw    (320,968) -- (325.86,968) -- (337.25,968) ;
\draw [shift={(339.25,968)}, rotate = 180] [color={rgb, 255:red, 0; green, 0; blue, 0 }  ][line width=0.75]    (4.37,-1.96) .. controls (2.78,-0.92) and (1.32,-0.27) .. (0,0) .. controls (1.32,0.27) and (2.78,0.92) .. (4.37,1.96)   ;
\draw [shift={(318,968)}, rotate = 0] [color={rgb, 255:red, 0; green, 0; blue, 0 }  ][line width=0.75]    (4.37,-1.96) .. controls (2.78,-0.92) and (1.32,-0.27) .. (0,0) .. controls (1.32,0.27) and (2.78,0.92) .. (4.37,1.96)   ;
%Straight Lines [id:da9314770721887407] 
\draw    (302.4,930) -- (288.42,930) ;
\draw [shift={(286.42,930)}, rotate = 360] [color={rgb, 255:red, 0; green, 0; blue, 0 }  ][line width=0.75]    (4.37,-1.96) .. controls (2.78,-0.92) and (1.32,-0.27) .. (0,0) .. controls (1.32,0.27) and (2.78,0.92) .. (4.37,1.96)   ;
\draw [shift={(304.4,930)}, rotate = 180] [color={rgb, 255:red, 0; green, 0; blue, 0 }  ][line width=0.75]    (4.37,-1.96) .. controls (2.78,-0.92) and (1.32,-0.27) .. (0,0) .. controls (1.32,0.27) and (2.78,0.92) .. (4.37,1.96)   ;
%Shape: Rectangle [id:dp058997893239281174] 
\draw  [fill={rgb, 255:red, 80; green, 227; blue, 194 }  ,fill opacity=0.36 ] (304.57,866) -- (356,866) -- (356,952) -- (304.57,952) -- cycle ;

% Text Node
\draw (384.13,969) node  [font=\footnotesize] [align=left] {\begin{minipage}[lt]{59.67pt}\setlength\topsep{0pt}
\begin{center}
{\small Deontic relation}
\end{center}

\end{minipage}};
% Text Node
\draw (286,969) node  [font=\footnotesize] [align=left] {\begin{minipage}[lt]{13.75pt}\setlength\topsep{0pt}
\begin{center}
{\small Set}
\end{center}

\end{minipage}};
% Text Node
\draw (330.5,851) node   [align=left] {$\displaystyle \mathcal{OS}$};
% Text Node
\draw (201,929) node   [align=left] {$\displaystyle \mathcal{SG}$};
% Text Node
\draw (331,875) node   [align=left] {$\displaystyle \mathcal{DS}$};
% Text Node
\draw (245.87,851) node   [align=left] {$\displaystyle \mathcal{SS}$};
% Text Node
\draw (239.38,881) node   [align=left] {$\displaystyle \mathcal{G}r$};
% Text Node
\draw (412.5,855) node   [align=left] {$\displaystyle \mathcal{FS}$};
% Text Node
\draw (412.04,879) node   [align=left] {$\displaystyle \mathcal{SCH}$};
% Text Node
\draw (276.38,929) node   [align=left] {$\displaystyle \mathcal{R}$};
% Text Node
\draw (412.99,927) node   [align=left] {$\displaystyle \mathcal{P}$};
% Text Node
\draw (441.01,927) node   [align=left] {$\displaystyle \mathcal{G}$};
% Text Node
\draw (431.7,902) node   [align=left] {$\displaystyle {nm}$};
% Text Node
\draw (395.21,901) node   [align=left] {$\displaystyle {mo}$};
% Text Node
\draw (381,855) node   [align=left] {$\displaystyle \mathcal{PO}$};
% Text Node
\draw (331,929) node   [align=left] {$\displaystyle \mathcal{OBL}$};
% Text Node
\draw (330,901) node   [align=left] {$\displaystyle \mathcal{PER}$};
% Text Node
\draw (278,851) node   [align=left] {$\displaystyle \mathcal{R}_{ss}$};
% Text Node
\draw (195,853) node   [align=left] {$\displaystyle \mathcal{IR}$};
% Text Node
\draw (253,929) node   [align=left] {$\displaystyle \mathnormal{ng}$};
% Text Node
\draw (383,927) node   [align=left] {$\displaystyle \mathcal{M}$};
% Text Node
\draw (272.51,902) node   [align=left] {$\displaystyle \mathcal{C}^{inter}$};
% Text Node
\draw (272.89,880) node   [align=left] {$\displaystyle \mathcal{C}^{intra}$};
% Text Node
\draw (207,880) node   [align=left] {$\displaystyle \mathcal{L}^{intra}$};
% Text Node
\draw (207.5,902) node   [align=left] {$\displaystyle \mathcal{L}^{inter}$};
% Text Node
\draw (227,929) node   [align=left] {$\displaystyle \mathnormal{np}$};

\end{tikzpicture}

%% file: figures/mm_synthesis_single_column.tex
\tikzset{every picture/.style={line width=0.75pt}} %set default line width to 0.75pt        

\begin{tikzpicture}[x=0.75pt,y=0.75pt,yscale=-1,xscale=1]
%uncomment if require: \path (0,2584); %set diagram left start at 0, and has height of 2584

%Straight Lines [id:da4973066741986565] 
\draw [line width=1.5]    (118.21,2302.58) -- (203.1,2302) ;
%Straight Lines [id:da14807114776731778] 
\draw    (368.35,2272) -- (368.35,2294) -- (332.16,2294) ;
\draw [shift={(330.16,2294)}, rotate = 360] [color={rgb, 255:red, 0; green, 0; blue, 0 }  ][line width=0.75]    (6.56,-1.97) .. controls (4.17,-0.84) and (1.99,-0.18) .. (0,0) .. controls (1.99,0.18) and (4.17,0.84) .. (6.56,1.97)   ;
%Straight Lines [id:da16285043353898754] 
\draw [line width=1.5]    (83.88,2398) -- (204.61,2398) ;
%Straight Lines [id:da6299512000169913] 
\draw    (169.94,2348) -- (204.61,2348) ;
%Straight Lines [id:da64750232417664] 
\draw    (84.65,2446) -- (383.15,2446) ;
%Straight Lines [id:da35895220906699743] 
\draw    (84.65,2230) -- (383,2230) ;
%Straight Lines [id:da715014372569708] 
\draw    (244.68,2262) -- (224.68,2262) -- (224.68,2408) ;
\draw [shift={(224.68,2410)}, rotate = 270] [color={rgb, 255:red, 0; green, 0; blue, 0 }  ][line width=0.75]    (6.56,-1.97) .. controls (4.17,-0.84) and (1.99,-0.18) .. (0,0) .. controls (1.99,0.18) and (4.17,0.84) .. (6.56,1.97)   ;
%Straight Lines [id:da71870438525014] 
\draw    (251.96,2328) -- (286.51,2328) -- (286.51,2356) ;
\draw [shift={(286.51,2358)}, rotate = 270] [color={rgb, 255:red, 0; green, 0; blue, 0 }  ][line width=0.75]    (6.56,-1.97) .. controls (4.17,-0.84) and (1.99,-0.18) .. (0,0) .. controls (1.99,0.18) and (4.17,0.84) .. (6.56,1.97)   ;
%Straight Lines [id:da6006267784187092] 
\draw [line width=0.75]  [dash pattern={on 0.84pt off 2.51pt}]  (252.87,2378) -- (252.87,2407) ;
\draw [shift={(252.87,2410)}, rotate = 270] [fill={rgb, 255:red, 0; green, 0; blue, 0 }  ][line width=0.08]  [draw opacity=0] (5.36,-2.57) -- (0,0) -- (5.36,2.57) -- cycle    ;
%Straight Lines [id:da8743336135156266] 
\draw    (322.88,2304) -- (322.88,2316) ;
\draw [shift={(322.88,2318)}, rotate = 270] [color={rgb, 255:red, 0; green, 0; blue, 0 }  ][line width=0.75]    (6.56,-1.97) .. controls (4.17,-0.84) and (1.99,-0.18) .. (0,0) .. controls (1.99,0.18) and (4.17,0.84) .. (6.56,1.97)   ;
%Straight Lines [id:da14641229967966152] 
\draw [line width=0.75]  [dash pattern={on 0.84pt off 2.51pt}]  (322.88,2378) -- (322.88,2407) ;
\draw [shift={(322.88,2410)}, rotate = 270] [fill={rgb, 255:red, 0; green, 0; blue, 0 }  ][line width=0.08]  [draw opacity=0] (5.36,-2.57) -- (0,0) -- (5.36,2.57) -- cycle    ;
%Straight Lines [id:da9260929933425808] 
\draw [line width=0.75]  [dash pattern={on 0.84pt off 2.51pt}]  (286.51,2378) -- (286.51,2420) -- (312.61,2420) ;
\draw [shift={(315.61,2420)}, rotate = 180] [fill={rgb, 255:red, 0; green, 0; blue, 0 }  ][line width=0.08]  [draw opacity=0] (5.36,-2.57) -- (0,0) -- (5.36,2.57) -- cycle    ;
%Straight Lines [id:da3057006030233673] 
\draw [line width=0.75]  [dash pattern={on 0.84pt off 2.51pt}]  (274,2449.7) -- (274,2457) ;
\draw [shift={(274,2460)}, rotate = 270] [fill={rgb, 255:red, 0; green, 0; blue, 0 }  ][line width=0.08]  [draw opacity=0] (3.57,-1.72) -- (0,0) -- (3.57,1.72) -- cycle    ;
%Straight Lines [id:da07288166228322246] 
\draw    (342,2449.98) -- (342,2458) ;
\draw [shift={(342,2460)}, rotate = 270] [color={rgb, 255:red, 0; green, 0; blue, 0 }  ][line width=0.75]    (6.56,-1.97) .. controls (4.17,-0.84) and (1.99,-0.18) .. (0,0) .. controls (1.99,0.18) and (4.17,0.84) .. (6.56,1.97)   ;
%Shape: Ellipse [id:dp8508274348425935] 
\draw   (95.09,2288.86) .. controls (95.09,2287.28) and (96.33,2286) .. (97.85,2286) .. controls (99.38,2286) and (100.62,2287.28) .. (100.62,2288.86) .. controls (100.62,2290.44) and (99.38,2291.71) .. (97.85,2291.71) .. controls (96.33,2291.71) and (95.09,2290.44) .. (95.09,2288.86) -- cycle ;
%Straight Lines [id:da3825450168053828] 
\draw    (97.85,2291.71) -- (97.85,2298.86) ;
%Straight Lines [id:da521321206042058] 
\draw    (97.85,2298.86) -- (93.71,2306) ;
%Straight Lines [id:da055514206493922025] 
\draw    (97.85,2298.86) -- (102,2306) ;
%Straight Lines [id:da8996496708356774] 
\draw    (102,2294.57) -- (93.71,2294.57) ;

%Straight Lines [id:da31678488015771755] 
\draw [line width=2.25]    (188,2454) -- (196.97,2454) ;
\draw [shift={(201.97,2454)}, rotate = 180] [fill={rgb, 255:red, 0; green, 0; blue, 0 }  ][line width=0.08]  [draw opacity=0] (5.72,-2.75) -- (0,0) -- (5.72,2.75) -- cycle    ;
%Shape: Ellipse [id:dp3927356466672782] 
\draw   (238.88,2451.17) .. controls (238.88,2450.36) and (239.67,2449.7) .. (240.64,2449.7) .. controls (241.61,2449.7) and (242.4,2450.36) .. (242.4,2451.17) .. controls (242.4,2451.99) and (241.61,2452.65) .. (240.64,2452.65) .. controls (239.67,2452.65) and (238.88,2451.99) .. (238.88,2451.17) -- cycle ;
%Straight Lines [id:da3365602555559104] 
\draw    (240.64,2452.65) -- (240.64,2456.32) ;
%Straight Lines [id:da7990875235744026] 
\draw    (240.64,2456.32) -- (238,2460) ;
%Straight Lines [id:da23945649338821617] 
\draw    (240.64,2456.32) -- (243.28,2460) ;
%Straight Lines [id:da11927353559661591] 
\draw    (243.28,2454.12) -- (238,2454.12) ;

%Straight Lines [id:da5816423191130675] 
\draw    (251.96,2272) -- (252.85,2356) ;
\draw [shift={(252.87,2358)}, rotate = 269.39] [color={rgb, 255:red, 0; green, 0; blue, 0 }  ][line width=0.75]    (6.56,-1.97) .. controls (4.17,-0.84) and (1.99,-0.18) .. (0,0) .. controls (1.99,0.18) and (4.17,0.84) .. (6.56,1.97)   ;
%Straight Lines [id:da9310455126832857] 
\draw    (321.97,2338) -- (321.97,2356) ;
\draw [shift={(321.97,2358)}, rotate = 270] [color={rgb, 255:red, 0; green, 0; blue, 0 }  ][line width=0.75]    (6.56,-1.97) .. controls (4.17,-0.84) and (1.99,-0.18) .. (0,0) .. controls (1.99,0.18) and (4.17,0.84) .. (6.56,1.97)   ;
%Shape: Rectangle [id:dp293492578719597] 
\draw   (120,2453) .. controls (120,2451.34) and (121.34,2450) .. (123,2450) -- (127.72,2450) .. controls (129.37,2450) and (130.72,2451.34) .. (130.72,2453) -- (130.72,2457) .. controls (130.72,2458.66) and (129.37,2460) .. (127.72,2460) -- (123,2460) .. controls (121.34,2460) and (120,2458.66) .. (120,2457) -- cycle ;
%Straight Lines [id:da33566712615128225] 
\draw    (261.05,2262) -- (306,2262) ;
\draw [shift={(308,2262)}, rotate = 180] [color={rgb, 255:red, 0; green, 0; blue, 0 }  ][line width=0.75]    (6.56,-1.97) .. controls (4.17,-0.84) and (1.99,-0.18) .. (0,0) .. controls (1.99,0.18) and (4.17,0.84) .. (6.56,1.97)   ;
%Shape: Rectangle [id:dp28383270948937667] 
\draw   (308,2236) -- (381.08,2236) -- (381.08,2276) -- (308,2276) -- cycle ;
%Straight Lines [id:da18020989903965012] 
\draw [line width=3]    (97.22,2282) -- (97.22,2270) ;
\draw [shift={(97.22,2264)}, rotate = 90] [fill={rgb, 255:red, 0; green, 0; blue, 0 }  ][line width=0.08]  [draw opacity=0] (10.18,-4.89) -- (0,0) -- (10.18,4.89) -- cycle    ;
%Straight Lines [id:da018421338049046554] 
\draw [line width=3]    (97.22,2310) -- (97.11,2322.37) ;
\draw [shift={(97.22,2328)}, rotate = 268.86] [fill={rgb, 255:red, 0; green, 0; blue, 0 }  ][line width=0.08]  [draw opacity=0] (10.18,-4.89) -- (0,0) -- (10.18,4.89) -- cycle    ;
%Shape: Rectangle [id:dp7281037051878541] 
\draw   (85.42,2450) -- (96.13,2450) -- (96.13,2460) -- (85.42,2460) -- cycle ;

% Text Node
\draw  [fill={rgb, 255:red, 255; green, 255; blue, 255 }  ,fill opacity=1 ]  (241.82,2323) .. controls (241.82,2320.24) and (244.06,2318) .. (246.82,2318) -- (259.82,2318) .. controls (262.58,2318) and (264.82,2320.24) .. (264.82,2323) -- (264.82,2333) .. controls (264.82,2335.76) and (262.58,2338) .. (259.82,2338) -- (246.82,2338) .. controls (244.06,2338) and (241.82,2335.76) .. (241.82,2333) -- cycle  ;
\draw (253.32,2328) node  [font=\scriptsize] [align=left] {$\displaystyle \boldsymbol{rcg}$};
% Text Node
\draw    (337,2252) -- (354,2252) -- (354,2272) -- (337,2272) -- cycle  ;
\draw (345.5,2262) node  [font=\scriptsize] [align=left] {$\displaystyle \mathcal{Y}$};
% Text Node
\draw    (311,2252) -- (334,2252) -- (334,2272) -- (311,2272) -- cycle  ;
\draw (322.5,2262) node  [font=\scriptsize] [align=left] {$\displaystyle \mathcal{T_{C}}$};
% Text Node
\draw    (357.39,2252) -- (378.39,2252) -- (378.39,2272) -- (357.39,2272) -- cycle  ;
\draw (367.89,2262) node  [font=\scriptsize] [align=left] {$\displaystyle \mathcal{M}$};
% Text Node
\draw (347.43,2244) node  [font=\scriptsize] [align=left] {$\displaystyle \mathcal{DS}$};
% Text Node
\draw  [fill={rgb, 255:red, 255; green, 255; blue, 255 }  ,fill opacity=1 ]  (270,2257) .. controls (270,2254.24) and (272.24,2252) .. (275,2252) -- (288,2252) .. controls (290.76,2252) and (293,2254.24) .. (293,2257) -- (293,2267) .. controls (293,2269.76) and (290.76,2272) .. (288,2272) -- (275,2272) .. controls (272.24,2272) and (270,2269.76) .. (270,2267) -- cycle  ;
\draw (281.5,2262) node  [font=\scriptsize] [align=left] {$\displaystyle \boldsymbol{rds}$};
% Text Node
\draw (158,2454.5) node  [font=\tiny] [align=left] {Relation name};
% Text Node
\draw (106.46,2454.5) node  [font=\tiny] [align=left] {Set};
% Text Node
\draw (255.47,2454.5) node  [font=\tiny] [align=left] {User};
% Text Node
\draw (216.32,2454.5) node  [font=\tiny] [align=left] {Define};
% Text Node
\draw (366.91,2454.5) node  [font=\tiny] [align=left] {Set relation};
% Text Node
\draw (306.61,2454.5) node  [font=\tiny] [align=left] {Impact on training};
% Text Node
\draw    (244.82,2410) -- (261.82,2410) -- (261.82,2430) -- (244.82,2430) -- cycle  ;
\draw (253.32,2420) node  [font=\scriptsize] [align=left] {$\displaystyle \boldsymbol{A}$};
% Text Node
\draw    (314.84,2410) -- (331.84,2410) -- (331.84,2430) -- (314.84,2430) -- cycle  ;
\draw (323.34,2420) node  [font=\scriptsize] [align=left] {$\displaystyle \boldsymbol{R}$};
% Text Node
\draw    (215.63,2410) -- (234.63,2410) -- (234.63,2430) -- (215.63,2430) -- cycle  ;
\draw (225.13,2420) node  [font=\scriptsize] [align=left] {$\displaystyle \mathcal{A}$};
% Text Node
\draw  [fill={rgb, 255:red, 255; green, 255; blue, 255 }  ,fill opacity=1 ]  (309.97,2363) .. controls (309.97,2360.24) and (312.21,2358) .. (314.97,2358) -- (328.97,2358) .. controls (331.73,2358) and (333.97,2360.24) .. (333.97,2363) -- (333.97,2373) .. controls (333.97,2375.76) and (331.73,2378) .. (328.97,2378) -- (314.97,2378) .. controls (312.21,2378) and (309.97,2375.76) .. (309.97,2373) -- cycle  ;
\draw (321.97,2368) node  [font=\scriptsize] [align=left] {$\displaystyle \boldsymbol{grg}$};
% Text Node
\draw    (274.56,2363) .. controls (274.56,2360.24) and (276.8,2358) .. (279.56,2358) -- (292.56,2358) .. controls (295.32,2358) and (297.56,2360.24) .. (297.56,2363) -- (297.56,2373) .. controls (297.56,2375.76) and (295.32,2378) .. (292.56,2378) -- (279.56,2378) .. controls (276.8,2378) and (274.56,2375.76) .. (274.56,2373) -- cycle  ;
\draw (286.06,2368) node  [font=\scriptsize] [align=left] {$\displaystyle \boldsymbol{rrg}$};
% Text Node
\draw    (240.87,2363) .. controls (240.87,2360.24) and (243.11,2358) .. (245.87,2358) -- (259.87,2358) .. controls (262.63,2358) and (264.87,2360.24) .. (264.87,2363) -- (264.87,2373) .. controls (264.87,2375.76) and (262.63,2378) .. (259.87,2378) -- (245.87,2378) .. controls (243.11,2378) and (240.87,2375.76) .. (240.87,2373) -- cycle  ;
\draw (252.87,2368) node  [font=\scriptsize] [align=left] {$\displaystyle \boldsymbol{rag}$};
% Text Node
\draw (156.15,2380.5) node  [font=\footnotesize] [align=left] {$\displaystyle  \begin{array}{{>{\displaystyle}l}}
\ \ \boldsymbol{Constraint\ Guides}\\
( roles\ and\ goals\ logic)
\end{array}$};
% Text Node
\draw  [fill={rgb, 255:red, 255; green, 255; blue, 255 }  ,fill opacity=1 ]  (309.93,2323) .. controls (309.93,2320.24) and (312.17,2318) .. (314.93,2318) -- (329.93,2318) .. controls (332.69,2318) and (334.93,2320.24) .. (334.93,2323) -- (334.93,2333) .. controls (334.93,2335.76) and (332.69,2338) .. (329.93,2338) -- (314.93,2338) .. controls (312.17,2338) and (309.93,2335.76) .. (309.93,2333) -- cycle  ;
\draw (322.43,2328) node  [font=\scriptsize] [align=left] {$\displaystyle \boldsymbol{gcg}$};
% Text Node
\draw  [fill={rgb, 255:red, 255; green, 255; blue, 255 }  ,fill opacity=1 ]  (214.77,2323) .. controls (214.77,2320.24) and (217.01,2318) .. (219.77,2318) -- (227.77,2318) .. controls (230.53,2318) and (232.77,2320.24) .. (232.77,2323) -- (232.77,2333) .. controls (232.77,2335.76) and (230.53,2338) .. (227.77,2338) -- (219.77,2338) .. controls (217.01,2338) and (214.77,2335.76) .. (214.77,2333) -- cycle  ;
\draw (223.77,2328) node  [font=\scriptsize] [align=left] {$\displaystyle \boldsymbol{ar}$};
% Text Node
\draw  [fill={rgb, 255:red, 255; green, 255; blue, 255 }  ,fill opacity=1 ]  (356.85,2289) .. controls (356.85,2286.24) and (359.08,2284) .. (361.85,2284) -- (374.85,2284) .. controls (377.61,2284) and (379.85,2286.24) .. (379.85,2289) -- (379.85,2299) .. controls (379.85,2301.76) and (377.61,2304) .. (374.85,2304) -- (361.85,2304) .. controls (359.08,2304) and (356.85,2301.76) .. (356.85,2299) -- cycle  ;
\draw (368.35,2294) node  [font=\scriptsize] [align=left] {$\displaystyle \boldsymbol{mo}$};
% Text Node
\draw (127,2344.5) node  [font=\small] [align=left] {$\displaystyle  \begin{array}{{>{\displaystyle}l}}
\mathbf{MOISE\ +MARL}\\
\mathbf{Specs.\ Level}
\end{array}$};
% Text Node
\draw (127,2422.5) node  [font=\small] [align=left] {$\displaystyle  \begin{array}{{>{\displaystyle}l}}
\mathbf{MARL\ Level}\\
\mathbf{( Dec-POMDP)}
\end{array}$};
% Text Node
\draw    (243.91,2252) -- (260.91,2252) -- (260.91,2272) -- (243.91,2272) -- cycle  ;
\draw (252.41,2262) node  [font=\scriptsize] [align=left] {$\displaystyle \mathcal{R}$};
% Text Node
\draw (182.64,2315) node  [font=\footnotesize] [align=left] {$\displaystyle \boldsymbol{Linkers}$};
% Text Node
\draw    (313.93,2284) -- (330.93,2284) -- (330.93,2304) -- (313.93,2304) -- cycle  ;
\draw (322.43,2294) node  [font=\scriptsize] [align=left] {$\displaystyle \mathcal{G}$};
% Text Node
\draw (127,2261.5) node  [font=\small] [align=left] {$\displaystyle  \begin{array}{{>{\displaystyle}l}}
\mathbf{{\displaystyle Org.\ Specs.\ Level}}\\
{\displaystyle \ \ \ \ \ \ \ \ \ \ \ (\mathcal{M}\mathbf{OISE^+})}
\end{array}$};

\end{tikzpicture}